\documentclass[a4paper,fleqn]{cas-sc}

\usepackage{hyperref}
\hypersetup{
    colorlinks,
    linkcolor=blue,
    filecolor=blue,      
    urlcolor=blue,
    citecolor=blue,
}

\usepackage{bbding}

\usepackage{graphicx}
\usepackage{subfigure}
\usepackage{CJKutf8}
\usepackage{algorithm}
\usepackage{algorithmic}
\usepackage{multirow}
\usepackage{makecell}
\usepackage[authoryear]{natbib}

\def\tsc#1{\csdef{#1}{\textsc{\lowercase{#1}}\xspace}}
\tsc{WGM}
\tsc{QE}
\tsc{EP}
\tsc{PMS}
\tsc{BEC}
\tsc{DE}

\begin{document}
\begin{CJK}{UTF8}{gbsn}
\let\WriteBookmarks\relax
\def\floatpagepagefraction{1}
\def\textpagefraction{.001}

\shorttitle{A Diversity Knowledge Enhanced MWP solver}

\shortauthors{Zhang et. al.}

\title [mode = title]{A Diversity-Enhanced Knowledge Distillation Model for Practical Math Word Problem Solving}                      

\affiliation[1]{organization={Faculty of Artificial Intelligence in Education, Central China Normal University},
            city={Wuhan},
            postcode={430079}, 
            state={Hubei},
            country={China}}
\affiliation[2]{organization={School of  Information and Safety Engineering, Zhongnan University of Economics and Law},
            city={Wuhan},
            postcode={430073}, 
            state={Hubei},
            country={China}}
\affiliation[3]{organization={Hubei Provincial Key Laboratory of Artificial Intelligence and Smart Learning \& School of Computer, Central China Normal University},
            city={Wuhan},
            postcode={430079}, 
            state={Hubei},
            country={China}}
\affiliation[4]{organization={Information Retrieval and Knowledge Management Research Lab, York University},
            city={Toronto},
            country={Canada}}


\author[1,2]{Yi Zhang}
\ead{yzhang@zuel.edu.cn}

\author[3]{Guangyou Zhou}
\ead{gyzhou@mail.ccnu.edu.cn}

\author[3]{Zhiwen Xie}
\ead{zwxie@ccnu.edu.cn}

\author[1]{Jinjin Ma}
\ead{majinjin@mails.ccnu.edu.cn}

\author[4]{Jimmy Xiangji Huang}
\ead{jhuang@yorku.ca}

\begin{abstract}
Math Word Problem (MWP) solving is a critical task in natural language processing, has garnered significant research interest in recent years. Various recent studies heavily rely on Seq2Seq models and their extensions (e.g., Seq2Tree and Graph2Tree) to generate mathematical equations. While effective, these models struggle to generate diverse but counterpart solution equations, limiting their generalization across various math problem scenarios. In this paper, we introduce a novel Diversity-enhanced Knowledge Distillation (DivKD) model for practical MWP solving. Our approach proposes an adaptive diversity distillation method, in which a student model learns diverse equations by selectively transferring high-quality knowledge from a teacher model. Additionally, we design a diversity prior-enhanced student model to better capture the diversity distribution of equations by incorporating a conditional variational auto-encoder. Extensive experiments on {four} MWP benchmark datasets demonstrate that our approach achieves higher answer accuracy than strong baselines while maintaining high efficiency for practical applications.


\end{abstract}




\maketitle
\section{Introduction}
The ability for mathematical reasoning has long been recognized as a fundamental challenge for computers \citep{Bobrow1964}. Math word problem (MWP) solving aims to generate solutions from mathematical problems expressed in natural language. Solving MWP requires natural language understanding and mathematical reasoning skills, which have garnered significant attention from the fields of natural language processing (NLP) and smart education. 

Researchers developed various methods to address MWP solving tasks, including statistical machine learning methods \citep{kushman2014learning,hosseini-etal-2014-learning,learning2016mitra,roy2018mapping}, semantic parsing methods \citep{shi2015automatically,koncel2015parsing,huang2017learning}, and deep learning methods \citep{wang2017DNS, xiao2023A, Zhang2024number}. Traditional statistical machine learning and semantic parsing methods require manually crafted feature engineering or template design, which are challenging to scale to large datasets. Recently, deep learning models have emerged as a promising paradigm for MWP solving. Leveraging the success of end-to-end models in NLP, researchers introduced Sequence-to-Sequence (Seq2Seq) models for MWP tasks \citep{wang2017DNS, Wang2019template,li2019modeling}. Seq2Seq models utilize an Encoder-to-Decoder (Enc2Dec) framework to translate the input problem description into a mathematical equation. However, these models often overlook the structural information of equations, potentially generating invalid equations that cannot be computed. To better capture equation structure, some studies have enhanced Seq2Seq models by incorporating tree-based decoders, such as Seq2Tree models (e.g., GTS) \citep{Xie19Goal_driven}) and Graph2Tree models (e.g., Graph2Tree-Z \citep{ZhangGraph2Tree2020}).

\begin{figure}[!t]
\centering
\includegraphics[width=0.75\textwidth]{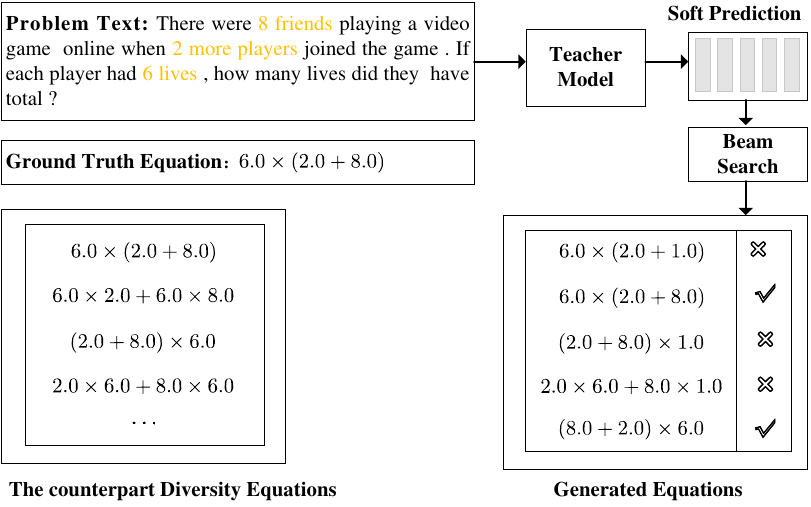}
\caption{An example of diversity equations and noise soft labels generated by the teacher model Graph2Tree-Z \citep{ZhangGraph2Tree2020}, in which the symbols ``$  \times $'' and ``$  \checkmark $'' indicate the error and true answer.}
\label{fig_example}
\end{figure}

Although existing Enc2Dec models (e.g., Seq2Seq, Seq2Tree and Graph2Tree) have achieved remarkable progress in solving MWPs, they are limited in modeling the diversity of solution equations due to data and model limitations. On the one hand, benchmark datasets (e.g., Math23K \citep{wang2017DNS} and MAWPS \citep{koncel2016mawps}) typically provide only one ground-truth solution equation for each math problem, even though numerous alternative correct equations can solve the same problem. For instance, in the math problem shown in Figure~\ref{fig_example}, the ground-truth equation is
``$6.0\times (8.0+2.0)$'', but other equations can also correctly solve this problem, such as ``$6.0\times 8.0 + 6.0\times 2.0$'' and ``$6.0\times (2.0+8.0)$''. Conventional Seq2Seq, Seq2Tree, and Graph2Tree models only utilize a single labeled ground-truth equation and a fixed decoder to generate equations, limiting their ability to produce diverse solution equations.

To alleviate these issues, \cite{WangTranMWP18,Wang2019template} employ an equation normalization approach to reduce the diversity of solution equations by imposing restrictions on the order of mathematical operators. However, the effectiveness of this approach has proven to be limited. \cite{zhang2020teacher} propose a knowledge distillation approach, called TSN-MD, which encourages the student model to learn diverse equations under the soft supervision of the teacher model by using multiple decoders. However, the use of multiple decoders is less adaptable due to the large and uncertain number of possible equations. Moreover, employing multiple decoders significantly increases inference time, making it impractical for real-world application scenarios. Additionally, some misleading knowledge from the teacher model can impair the student model's performance. As the example shown in Figure~\ref{fig_example}, the teacher model generates an incorrect solution equation ranked first in the beam search results. This indicates that the soft prediction of the teacher model follows an inaccurate distribution. Consequently, teaching the student to learn these soft predictions can negatively impact the student model.

Therefore, considering these challenges described above, we propose a novel Diversity-Enhanced Knowledge Distillation (DivKD) model for practical MWP solving. Our approach introduces an Adaptive Diversity Knowledge Distillation (AdaKD) method to effectively transfer the teacher's diverse knowledge by selectively training the student model with high-quality soft and hard labels. This method adaptively selects the high-quality teacher labels as the diversity target labels for student model learning, and estimates the quality of the intermediate soft labels for training student discernment. The purpose of method is to address the scarcity of diverse solution equations in datasets. Secondly, we design a diversity prior enhanced student model, enabling the model to capture equation diversity by incorporating a Conditional Variational Autoencoder (CVAE) module. In this way, we can model the diversity distribution over possible solution equations by sampling latent variables via a diversity prior network. Extensive experiments conducted on four widely used benchmark MWP datasets demonstrate the effectiveness of our proposed method.

In summary, our main contributions are listed as follows:





\begin{itemize}
    \item We propose a novel Adaptive Diversity Knowledge Distillation (AdaKD) method, which selectively transfers high-quality knowledge from the pre-trained teacher model to better guide the learning of the student model. This method addresses the scarcity of diverse solution equations in datasets, enabling the student model to acquire both diverse and high-quality information from the pre-trained teacher model.
    
    \item We introduce a diversity prior-enhanced student model by incorporating Conditional Variational Autoencoder (CVAE) into existing models, enabling the model to capture the diversity distribution of solution equations. This approach leverages a diversity prior network to sample latent variables, thereby modeling the distribution of possible solution equations and generating diverse solution equations during the testing phase.
    
    \item We conducted extensive experiments on four MWP datasets to validate the effectiveness of our proposed Diversity-Enhanced Knowledge Distillation (DivKD) model. The experimental results show that, based on the answer accuracy metric, the proposed methods effectively improve the performance of existing models without sacrificing model efficiency.
\end{itemize}

The remainder of this paper is organized as follows: Section \ref{relation} introduces the related work of this study. Section \ref{preliminaries} provides background information and formally defines the research task. Section \ref{approach} presents our proposed DivKD approach. Section \ref{experiment} details the experimental results. Finally, Section \ref{conclusion} concludes the paper and discusses prospects for future work.

\section{Related Work}
\label{relation}

In automatically solving MWPs, deep learning techniques have become the primary approach for MWP solving due to their superior performance and generalization capabilities compared to traditional rule-based and statistical methods \citep{lu2023survey}. Recently, the deep learning models \citep{Xie19Goal_driven, ZhangGraph2Tree2020, Zhang2022multi, Bin2023solve, Qin2023Pretain, Bin2023non} for MWP solving have primarily focused on end-to-end frameworks (e.g., Seq2Seq, Seq2Tree, Graph2Tree). Besides, some subsequent methods, such as PARAMAWPS \citep{Raiyan2023math}, DiverseMWP \citep{Zhou2023learning} and MathEncoder \citep{Qin2023Pretain}, leverage data augmentation, voting, or task-specific pre-training strategies to enhance the performance. However, these approaches are not the focus of this paper due to differing motivations and research goals.

In addition to the aforementioned base solvers, recent pre-trained language models (PLMs) offer opportunity for developing more powerful MWP solvers, including MWP-BERT \citep{liang2022mwp}, Generate\&Rank \citep{shen2021generate} and Deductive Reasoner \citep{Learning2022Jie}. Most recently, the large-scale models have demonstrated remarkable reasoning abilities for MWP solving \citep{Shao2022chaining, Pi2022reasoning, Liang2023let}. Large language models (LLMs) achieve higher accuracy and explain the solution processes \citep{Touvron2023llama, GPT-3}. Large Multimodal models (LMMs) effectively incorporate visual information to aid mathematical reasoning \citep{mathllava2024, mathpuma2024}. However, the computational efficiency makes large size models hard for real-world situations especially in smart eduction domain where cost and speed are important consideration. Additionally, LLMs such as GPT-3 \citep{GPT-3} are prone to factual errors \citep{Ouyang2022training}, their knowledge bases are not up-to-date. In summary, a computationally efficient solver still needs to be developed to make it practical for real-world situations.

Although the previous MWP solvers have achieved promising performance, they are trained on datasets that provide only a single ground-truth target, which limits the solver's generalization ability. Consequently, some works attempt to construct rich template sketches to accommodate diverse equations during the training process \citep{learning2016mitra, huang2017learning, Wang2019template}. However, this approach can lead to many different equations with various combinations, resulting in a larger non-deterministic space during the decoding process. \cite{zhang2020teacher} propose to extract knowledge from a pre-trained teacher network and encourage multiple student networks to mimic the soft labels generated by the teacher network to learn diverse solutions, named TSN-MD. Subsequent paper \citep{wang2022structure} following TSN-MD focus on unifying the output solution structures through a proposed information storage structure called M-Tree. However, multiple teacher networks or solution structures also require additional model complexity to assess the data quality in mathematical reasoning.

Different from the above-mentioned models, we propose a new framework to better capture the diversity of solution equations in MWPs. Firstly, we design an AdaKD method to select high-quality distillation labels, enriching diversity by incorporating new knowledge from teacher models while reducing the impact of noise information. Secondly, we introduce an enhanced student model with a diversity prior, enabling the model to capture the diversity distribution of solution equations.

\section{Background}
\label{preliminaries}

In this section, we introduce the Knowledge Distillation method and the representative teacher-student framework, discuss the relationship in some preliminaries of this work, and finally, summarize the mathematical notations and definitions used in this study in Table~\ref{tab:notations}.

\subsection{Knowledge Distillation}
Knowledge distillation (KD) \citep{Hinton2015} has been widely used in deep learning models due to its ability to transfer knowledge from a pre-trained teacher model to a student model.  In the process of KD, a teacher model $T$ is pre-trained to generate soft targets, and then a student model $S$ is trained under the supervision of both the ground-truth labels and generated soft labels. Formally, the student model $S$ is trained on a linear combination of two loss functions:
\begin{equation}
    \mathcal{L}=(1-\lambda)\mathcal{L}_{CE}+\lambda \mathcal{L}_{KD},
\end{equation}
where $\lambda$ is a hyper-parameter, $\mathcal{L}_{CE}$ is the cross-entropy between the student output $S(x)$ and the ground-truth label $y$, which is computed as:
\begin{equation}
    \mathcal{L}_{CE}=-y\log\sigma(S(x)).
\end{equation}
where $\sigma$ denotes the softmax function. And $\mathcal{L}_{KD}$ is the Kullback-Leibler (KL) divergence loss between student output $S(x)$ and the soft target $T(x)$ generated from teacher model, namely:
\begin{equation}
    \mathcal{L}_{KD}=KL(\sigma(S(x)/\tau||\sigma(T(x)/\tau)))
\end{equation}
where $\tau$ is a temperature hyper-parameter in softmax function.

\subsection{Problem Definition}
Let $x=\{w_1,w_2,\cdots,w_n\}$ denote a math word problem and $y=\{a_1,a_2,\cdots,a_m\}$ denote the output solution equation sequence, where $w_i$ in the math word problem is a word token, $a_i$ in the answer equation is a numeric value or a operator, $n$ is the number of words in $x$ and $m$ is the number of elements in $y$. 
Given a set of MWPs and corresponding solution equations $D=\{(x,y)\}$, the task of solving math word problems aims to learn a model to map the text sequence of a given math word problem $x$ into an output equation sequence $y$.

\begin{table}[]\small
\centering 
\caption{Notations and Meanings.}
\label{tab:notations}
\begin{tabular}{|l|l||l|l|}
\hline
\textbf{Notation} & \textbf{Meanings}                & \textbf{Notation}  &   \textbf{Meanings}     \\
\hline
$x$      & An MWP textual description.     & $y$       & The target mathematical equation.    \\ \hline
$T$    & The teacher model in KD.   & $S$         & The student model in KD.    \\ \hline
$T(x)$    & The soft output of teacher model.   & $S(x)$   & The soft output of student model.    \\ \hline
$Encoder(\cdot)$    & One of the existing encoder models.   & $TreeDecoder(\cdot)$       & A Tree-based decoder.   \\ \hline
$h$    & The output of $Encoder(x)$.     & $\theta_T$    &  The parameters of the teacher model.   \\ \hline
$h_y$    &  Representation of target equation.     & $\theta_S$    &  The parameters of the student model.   \\ \hline
$h_z$    &  The representation of z.  & $z$    &  Latent distribution over possible solutions.  \\ \hline
$\mathcal{N}(\mu,\sigma^2\mathbf{I})$    &  Posterior distribution.     & $\mathcal{N}(\mu',\sigma'^2\mathbf{I})$    &  Diversity prior distribution .    \\ \hline
$\mu, \log \sigma^2$ & Gaussian parameters for $\mathcal{N}(\mu,\sigma^2\mathbf{I})$.    & $\mu', \log \sigma'^2$  & Gaussian parameters for $\mathcal{N}(\mu',\sigma'^2\mathbf{I})$.   \\ \hline
$W_\mu,W_\sigma,b_\mu,b_\sigma$ & Trainable parameters for $\{\mu, \log \sigma^2\}$.    & $W_{\mu'},W_{\sigma'},b_{\mu'},b_{\sigma'}$ & Trainable parameters for $\{\mu', \log \sigma'^2\}$.   \\ \hline
$\mathcal{L}_{CVAE}$ & The loss function of CVAE. & $\mathcal{L}_{AdaHardKD}$ & The loss function of adaptive hard KD.  \\ \hline
$w_x$ & Weight score of teacher’s soft labels. & $\mathcal{L}_{AdaSoftKD}$ & The loss function of adaptive soft KD.  \\ \hline
\end{tabular}
\end{table}

\begin{figure}[!t]
\centering
\includegraphics[width=0.98\textwidth]{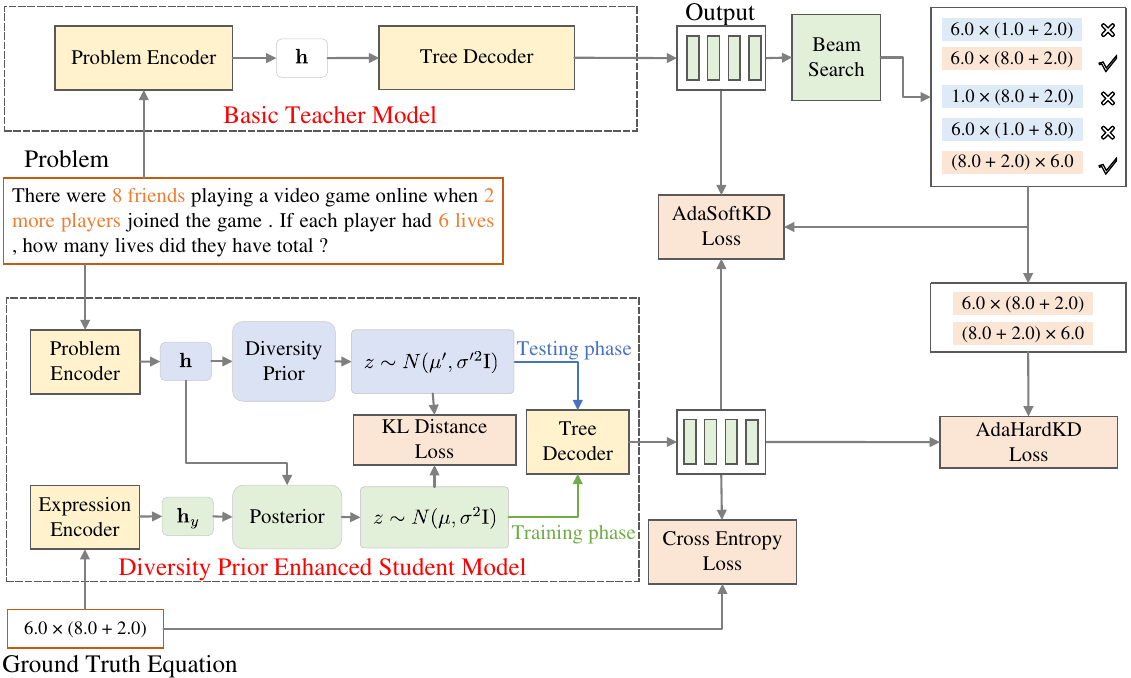}
\caption{The overview of the proposed DivKD model.}
\label{fig_model}
\end{figure}

\section{Our Method}
\label{approach}
Figure~\ref{fig_model} illustrates the overview of the proposed a diversity-enhanced knowledge distillation (DivKD) model.
In our approach, we model the diversity of MWPs using a knowledge distillation framework with a diversity prior.
Firstly, a diversity enhanced teacher-student framework is applied, where a basic teacher is trained to generate new knowledge and a diversity prior enhanced student model is proposed to model the diversity distribution of solution equations. 
Then, we propose two adaptive knowledge distillation strategies (e.g., adaptive soft distillation and hard distillation) to transfer high-quality knowledge from the teacher model to the student model.

\subsection{Diversity Enhanced Knowledge Distillation Framework}
In the task of MWP solving, there can be multiple correct solution equations for a given problem, while only one solution equation is annotated in datasets. Most previous models are designed to fit the single annotation solution equation without considering the diversity of solutions. Thanks to the generalization ability of deep learning models, we can leverage KD to learn knowledge beyond the original dataset from a teacher model. The KD framework consists of a teacher model and a student model. In this paper, we select tree-based encoder-decoder models (e.g., GTS \citep{Xie19Goal_driven} and Graph2Tree-Z \citep{ZhangGraph2Tree2020}) as basic structures since these models are typical in MWP solving task.  GTS \citep{Xie19Goal_driven} applies a bidirectional gated recurrent unit (BiGRU) as the encoder. Graph2Tree-Z \citep{ZhangGraph2Tree2020} uses a graph-based encoder (GraphEncoder) to better learn relationships and order information among quantities. Furthermore, we also present a PLM-enhanced version of the Graph2Tree-Z model named Ro-Graph2Tree-Z, which employs a pre-trained language (PLM) model (e.g., RoBERT \citep{RoBERTa2019}) as an encoder to enhance the understanding of MWPs. It's worth noting that our method is generic and can also be applied to other MWP solvers. 
In this subsection, we describe the structure of teacher and student models used in our KD framework.

\subsubsection{Basic Teacher Model}
Firstly, we use the original structure of the encoder-decoder model (e.g., GTS, Graph2Tree-Z and Ro-Graph2Tree-Z) as a teacher model to pre-train on MWP datasets.  
Given a math problem $x=\{w_1,w_2,\cdots,w_n\}$, the teacher model first uses an encoder to model the sequence of input and return the hidden state of the input math problem, which is denoted as:
\begin{equation}
    h=Encoder(x)
\end{equation}
where $h$ is the output hidden state of the encoder, $Encoder \in \{BiGRU, GraphEncoder, RoBERTaEncoder\}$ is one of the existing text encoder models, and BiGRU, GraphEncoder and RoBertaEncoder are the encoders in GTS, Graph2Tree-Z and Ro-Graph2Tree-Z, respectively.

Finally, the output feature of the encoder is fed into a tree-based decoder to generate the distribution of solution equations, which is denoted as:
\begin{equation}
    p(y|x,\theta_T)=TreeDecoder(h)
\end{equation}
where $\theta_T$ denotes the parameters of the teacher model.

\subsubsection{Diversity Prior Enhanced Student Model}
The variational autoencoder (VAE) \citep{KingmaW13, RezendeMW14} and its variants, such as conditional variational autoencoder (CVAE) \citep{SohnLY15}, are widely adopted generative models due to their ability of generating diverse data \citep{ZhaoZE17, WuWW20}. 
To model the diversity of solution equations, we introduce a diversity prior to enhancing the student model by incorporating a CVAE component into the original encoder-decoder MWP models (e.g., GTS, Graph2Tree-Z, Ro-Graph2Tree-Z). 
In the task of MWP solving, the generation of a solution equation $y$ is conditioned on a given math problem $x$, which can be denoted as $p(y|x)$. Using CVAE, a latent variable $z$ is introduced to capture the latent distribution over possible solution equations and the conditional likelihood calculation is reformulated as:
\begin{equation}
    p(y|x)=\int p(y|x,z)p(z|x)
\end{equation}\
Here $p(z|x)$ is a diversity prior distribution which is used to sample diverse latent variable $z$, and $p(y|x,z)$ is a generative distribution. 
Thus, the process of generating a solution equation is as follows: (1) sample a latent variable $z$ from the prior distribution $p(z|x)$ and (2) generate solution equation $y$ from the conditional generative distribution $p(y|x,z)$. 

It is intractable to directly optimize the conditional log-likelihood of $p(y|x)$ due to the marginalization over the latent variable $z$. To address this problem, a stochastic gradient variational bayes (SGVB) framework \citep{KingmaW13, SohnLY15} is applied to train the CAVE model, which converts the original conditional log-likelihood into variational lower bound:
\begin{equation}
    \mathcal{L}_{CVAE}=-KL(q(z|x,y)||p(z|x))+\mathbb{E}_{q(z|x,y)}\left[ \log p(y|x,z) \right] \leq \log p(y|x)
\end{equation}
where $q(z|x,y)$ is the posterior distribution. The first term is the KL distance between posterior and prior distribution, and the second term is the cross-entropy between output distribution and ground-truth labels. 

Following previous works \citep{KingmaW13, ZhaoZE17, ZhangDJH2016}, we assume that $z$ follows a multivariate Gaussian distribution, namely $q(z|x,y)\sim \mathcal{N}(\mu,\sigma^2\mathbf{I})$ and $p(z|x)\sim \mathcal{N}(\mu',\sigma'^2\mathbf{I})$. 
And we use two neural networks to approximate the diversity prior distribution $p(z|x)$ and posterior distribution $q(z|x,y)$. For the posterior distribution $q(z|x,y)$, we first apply a BiGRU network to encode the target equation to obtain the output representation:
\begin{equation}
  h_y = BiGRU(y)
\end{equation}
Then, linear feed-forward networks are applied to obtain the Gaussian parameters $\mu$ and $\log \sigma^2$ for the posterior distribution:
\begin{equation}
\begin{aligned}
     &\mu = W_\mu [h||h_y]+b_\mu \\
     &\log \sigma^2=W_\sigma [h||h_y]+b_\sigma \\
\end{aligned}
\end{equation}
where $W_\mu$, $W_\sigma$, $b_\mu$ and $b_\sigma$ are trainable parameters in neural networks. Similarly, we can compute the Gaussian parameters $\mu'$ and $\log \sigma'^2$ for the prior distribution using a linear prior network as follows:
\begin{equation}
\begin{aligned}
    &\mu' = W_{\mu'} h+b_{\mu'} \\
    &\log \sigma'^2=W_{\sigma'} h+b_{\sigma'} \\
\end{aligned}
\end{equation}

During training, we sample $z$ from posterior distribution $\mathcal{N}(\mu,\sigma^2\mathbf{I})$, while the latent variable $z$ is sampled from diversity prior distribution $\mathcal{N}(\mu',\sigma'^2\mathbf{I})$ during testing. 
Here, a reparameterization technique \citep{KingmaW13} is applied to obtain the representation of $z$, namely $h_z=\mu+\sigma\odot \epsilon, \epsilon\sim \mathcal{N}(0,\mathbf{I})$. 
Then, the latent variable can be used to generate solutions via tree-based decoder:
\begin{equation}
    p(y|x,\theta_S)=TreeDecoder(h+h_z)
\end{equation}
where $\theta_S$ is the parameters of the student model.

\subsection{Adaptive Diversity Knowledge Distillation}
Given that datasets typically provide only a single equation for each math problem,  it becomes challenging for a model to gain a comprehensive understanding of diverse solution equations for an MWP. 
In this study, we employ the KD method to address this issue. Previous KD-based approaches, such as TSN-MD, have solely focused on transferring soft labels from a teacher model to a student model without assessing the quality of the soft labels generated by the teacher. In practice, not all soft labels from the teacher model may enhance the student's performance. Some soft labels may contain noise information that can potentially degrade the student's performance. Therefore, how to measure the quality of teacher knowledge is of great importance to learning a good student model. A simple solution is to measure the quality of soft labels according to the cross-entropy of the teacher's prediction \citep{Li2021dynamic, Wang2021selective}. 
However, this method is not well-suited for MWP tasks because there may be multiple correct equations for a single MWP. If we simply employ cross-entropy between soft labels and labelled ground-truth equations, it may result in misjudgment for other unlabeled correct equations. 
To address this problem, we propose an adaptive diversity knowledge distillation (AdaKD) method for better distillation, which consists of two adaptive KD strategies, namely adaptive hard KD (AdaHardKD) and adaptive soft KD (AdaSoftKD).  Specifically, we introduce a beam search-based approach to assess the quality of the teacher's predictions. When provided with the soft predictions from the teacher model, we employ beam search to generate the top-$K$ equations. Subsequently, we calculate the results of these equations and compare them with the ground-truth answer value to determine their correctness. Intuitively, if an equation's result matches the ground-truth value, it is considered a correct equation. Thus, we can obtain a set of correct equations for MWPs, which is denoted as $D^{kd}=\{(x,y^{kd})\}$. These generated equations are viewed as new hard labels to compute the loss of adaptive hard KD, which is calculated as:
\begin{equation}
    \mathcal{L}_{AdaHardKD}=\frac{1}{|D^{kd}|}\sum_{(x,y^{kd})\in D^{kd}} -\log p(y^{kd}|x,\theta_S)
\end{equation}

In addition to using hard labels, we also introduce an adaptive soft KD strategy to learn high-quality knowledge from the soft labels of the teacher.  
Intuitively, a better distribution will yield a greater number of correct equations with higher ranks. 
Consequently, we can assess the quality of the teacher's soft labels based on the outcomes of the beam search. 
Formally, let $\mathcal{B}=\{(y^{kd}_k,r_k)|1\leq k\leq K\}$ denote the results of beam search, where $y^{kd}_k$ is the $k$-th equation and $r_k$ is its rank in the beam search results. And let $\mathcal{B}' \in \mathcal{B}$ denote the set of correct equations \footnote{To efficiently capture diverse knowledge of expression, we employ answer accuracy rather than expression accuracy as our evaluation metric in this paper. The equation is considered correct in the beam generated by the teacher model when its calculation value equals the ground truth.}. Then, the weight score of the teacher's soft prediction can be computed as:
\begin{equation}
    \omega_x=\frac{1}{K}\sum_{(y^{kd},r) \in \mathcal{B}'} \lambda^{r}
\end{equation}
where $\lambda\in (0,1]$ is an attenuation factor used to assign higher scores to top-ranked equations. 
By utilizing the weighted scores, we encourage the student model to gain more knowledge from high-quality soft labels. 
And we define the AdaSoftKD loss function as:
\begin{equation}
    \mathcal{L}_{AdaSoftKD}=\frac{1}{|D|}\sum_{(x,y)\in D} \omega_x KL(p(y|x,\theta_S)||q(y|x,\theta_T))
\end{equation}

\subsection{Training Objective}
Finally, the model is trained by using a linear combination of three losses, namely the CVAE loss, the adaptive soft KD loss and the adaptive hard KD loss:
\begin{equation}
    \mathcal{L}=-\mathcal{L}_{CVAE}+\beta \mathcal{L}_{AdaHardKD} +\gamma \mathcal{L}_{AdaSoftKD}
\end{equation}
where $\beta$ and $\gamma$ are hyper-parameters controlling the weights of $\mathcal{L}_{AdaHardKD}$ and $ \mathcal{L}_{AdaSoftKD}$. To summarize our DivKD method, we provide the algorithm pseudocode in Algorithm \ref{alg:DivKD}.

\begin{algorithm}[t]
\caption{Diversity-Enhanced Knowledge Distillation}
\label{alg:DivKD}
\begin{algorithmic}[1]
\STATE \textbf{Input:} Training dataset $\mathcal{D} = \{(x_i, y_i)\}_{i=1}^N$, teacher model $\mathcal{T}$, student model $\mathcal{S}$, number of beams $K$
\STATE \textbf{Output:} Trained student model $\mathcal{S}$
\STATE Initialize teacher model $\mathcal{T}$ and student model $\mathcal{S}$
\STATE Pre-train teacher model $\mathcal{T}$ on $\mathcal{D}$
\FOR{each batch $(x, y) \in \mathcal{D}$}
    \STATE Obtain teacher predictions $p(y|x, \theta_T) = \mathcal{T}(x)$
    \STATE Perform beam search to generate top-$K$ equations $\mathcal{B} = \{(y_k, r_k)\}_{k=1}^K$
    \STATE Identify correct equations $\mathcal{B}' \subseteq \mathcal{B}$
    
    \STATE \textbf{Calculate adaptive hard KD loss:}
    \STATE \quad $D^{kd} \gets \{(x, y^{kd}) \,|\, y^{kd} \in \mathcal{B}' \}$
    \STATE \quad $\mathcal{L}_{AdaHardKD} \gets -\frac{1}{|D^{kd}|} \sum_{(x, y^{kd}) \in D^{kd}} \log p(y^{kd}|x, \theta_S)$
    
    \STATE \textbf{Calculate adaptive soft KD loss:}
    \STATE \quad Compute weight score $\omega_x$ based on ranks in $\mathcal{B}$
    \STATE \quad $\mathcal{L}_{AdaSoftKD} \gets \frac{1}{|\mathcal{D}|} \sum_{(x,y) \in \mathcal{D}} \omega_x \cdot KL(p(y|x, \theta_S) || q(y|x, \theta_T))$
    
    \STATE \textbf{Total loss:}
    \STATE \quad $\mathcal{L}_{total} = -\mathcal{L}_{CVAE} + \beta \mathcal{L}_{AdaHardKD} + \gamma \mathcal{L}_{AdaSoftKD}$
    
    \STATE Update student model $\mathcal{S}$ using optimizer on $\mathcal{L}_{total}$
\ENDFOR
\STATE \textbf{Return} trained student model $\mathcal{S}$
\end{algorithmic}
\end{algorithm}

\section{Experiments}
\label{experiment}

In this section, we provide a comprehensive description of the datasets employed, the baseline methodologies applied, and the overall experimental settings. Subsequently, we present detailed findings from comparison and ablation experiments, which substantiate the efficacy of our approach. Our focus is directed towards investigating the following five research questions:

\begin{itemize}
    \item \textbf{RQ1}: How does our model perform on the answer accuracy metrics compared to existing MWP solvers?
    
    \item \textbf{RQ2}: How does our diversity-enhanced knowledge distillation approach, DivKD, perform compared to existing knowledge distillation methods in the MWP task?
    
    \item \textbf{RQ3}: How does the proposed adaptively diverse knowledge distillation (AdaKD) alleviate the limitations of single labeled ground-truth expression in the dataset?
    
    \item \textbf{RQ4}: How can our approach improve the performance of existing models without sacrificing model efficiency, especially over other KD methods?
    
    \item \textbf{RQ5}: How does our approach perform on real-world datasets, particularly in capturing implicit diversity knowledge from problem texts and expressions and alleviating the limitations of single-labeled ground-truth expressions in the datasets?
\end{itemize}

\subsection{General settings}
\subsubsection{Datasets}

\begin{table}[htbp]
  \centering
  \caption{Dataset statistics. }\label{data:details}
    \begin{tabular}{l|cccccc}
    \hline
    \textbf{Dataset} & \#Train & \#Dev   & \#Test  & \begin{tabular}[c]{@{}c@{}}\#Avg.\\ Problem Length\end{tabular} & \begin{tabular}[c]{@{}c@{}}\#Avg. \\ Operators\end{tabular}  & Language \\
    \hline
    \textbf{Math23K} & 21,161 & 1,000 &1,000 & 28.04 & 2.70   & Chinese \\
    \textbf{MAWPS}  & 1,589  & 199   & 199   & 30.32 & 1.61  & English \\
    \textbf{MathQA} & 16,191 & 2,411 & 1,605 & 39.63 & 4.63  & English \\
    \textbf{SVAMP} & 3,138   & -      & 1,000 & 31.87 & 1.33  & English \\
    \hline
    \end{tabular}%
\end{table}%

In our experiments, we validate our proposed method on four widely used MWP benchmarks: \textbf{Math23K} \citep{wang2017DNS}, \textbf{MAWPS} \citep{koncel2016mawps}, \textbf{MathQA} \citep{mathqa2019} and \textbf{SVAMP} \citep{Are2021Patel}. The overall dataset statistics are presented in Table \ref{data:details}.

Math23K is a large-scale Chinese dataset that consists of 23,161 instances. Each instance is annotated with only one ground-truth equation and its corresponding answer. Following previous studies \citep{ZhangGraph2Tree2020,zhang2020teacher}, we adopt the same data splits: 21,161 math problems for training, 1,000 instances for validation, and 1,000 instances for testing. We report results based on answer accuracy on the test data as well as the 5-fold average results. 

MAWPS is a standard English MWP-solving dataset that contains 1,987 instances. Since its small size and lack of standard data splits, we perform 5-fold cross-validation and report the average results on this dataset. 

MathQA is a large-scale English dataset designed to evaluate the performance of models in solving more complex mathematical problems, where each problem requires multiple operators to derive the final answer.

SVAMP is a challenging English benchmark derived from instances sampled from the MAWPS dataset, consisting of 3,138 training instances and 1,000 test instances. It introduces variations such as noun phrase exchanges and the addition of extra quantities to assess whether NLP models can effectively understand and interpret contextual information.

\subsubsection{Evaluation Metric} 

\textit{Answer accuracy} and \textit{expression accuracy} are two wildly used evaluation metrics in MWP task. \textit{Answer accuracy} indicates that a prediction is correct if the calculated value of the generated expression matches the ground-truth answer. \textit{Expression accuracy} assesses whether the structure of the generated equation is identical to the target expression. The former measures the correctness of the answer, while the latter is concerned with the correctness of the structure. In our paper, we follow most of the prior works \citep{ZhangGraph2Tree2020, Learning2022Jie, Bin2023non} employing the \textit{answer accuracy} as our standard evaluation metrics due to our focus on diversity knowledge inherent in datasets and teacher predictions to enhance the student model's ability to generate diverse equations. \textit{Answer accuracy} highlights the potential of DivKD in diversity-driven mathematical problem-solving.

\begin{table}[htbp]
  \centering
  \caption{The search scopes and the hyper-parameter setting.}
    \begin{tabular}{l|ccccc}
    \toprule
    \textbf{Number of Parameter} & \textbf{Search Scope} & \textbf{Math23K} & \textbf{MAWPS} & \textbf{MathQA} & \textbf{SVAMP}\\
    \hline
    \#epoch & \{80,100,120,150\} & 80    & 80    & 150 & 80 \\
    batch size & \{8,16,30,64\} & 30    & 30    & 16  &30 \\
    RoBERTa's learning rate & \{2e-5,5e-5,5e-6\} & 5e-5  & 5e-5  & 2e-5 & 5e-5\\
    network layer learning rate & \{5e-4,1e-3,5e-3\} & 1e-3  & 1e-3  & 1e-3  & 1e-3 \\
    hidden size & \{384,512,768\} & 512   & 512   & 512 & 512\\
    beam search & \{3,5,7\} & 5     & 5     & 5 & 5\\
    attenuation factor & \{0.3,0.5,0.8\} & 0.8   & 0.8   & 0.8 & 0.8\\
    the weight of AdaHardKD & \{0.1,0.2,0.3,0.4\} & 0.3   & 0.1   & 0.1  & 0.1 \\
    the weight of AdaSoftKD & \{0.01,0.05,0.1,0.2\} & 0.1   & 0.05  & 0.05 & 0.05 \\
    \bottomrule
    \end{tabular}%
  \label{tab:param}%
\end{table}%

\subsubsection{Experimental Setting}
We implement our model using Pytorch \footnote{https://pytorch.org/} and conduct experiments on Ubuntu 20.04 using a server equipped with two NVIDIA RTX A6000 GPUs, each with 48 GB of GPU memory. The Adam optimizer is used in our model, and we use a grid search to select the appropriate hyperparameters based on the evaluation metrics on the test set. For the basic models such as GTS and Graph2Tree-Z, we set the initial learning rate to 1e-3, with the rate halving every 20 epochs. The word embedding dimension is set to 128, and the hidden state dimension to 512. For the basic models with pre-trained language models, such as Ro-Graph2Tree-Z, we use RoBerta-base \citep{RoBERTa2019}, setting the initial learning rate to 5e-5. For LLaMA \citep{Touvron2023llama}, we freeze its parameters and use it as an encoder for the input text. To generate more diverse equations, we set the beam search size to 5. The attenuation coefficient $\lambda$ is set to $0.8$, and the weight $\beta$ for AdaHardKD is set 0.3, 0.1, 0.1 and 0.1 for Math23K, MAWPS, MathQA and SVAMP, respectively. The weight $\gamma$ for AdaSoftKD is set to 0.1, 0.05, 0.05, 0.05 for the same datasets, respectively. 
Table \ref{tab:param} lists the main hyperparameters setting in the model and details of their search ranges. Our code and the model's detail are released at \textcolor{blue}{https://github.com/a773938364/DivKD}.

\subsubsection{Comparison Models}

Based on the architectures employed for math text encoding, we categorize the baseline models into three groups: \textbf{Group A}, which includes models utilizing small deep learning architectures such as LSTM, RNN, and GRU; \textbf{Group B} comprising models leveraging PLMs like BERT and RoBERTa; and \textbf{Group C}, consisting of models incorporating LLMs such as T5 and LLaMA.

\begin{itemize}
    \item \textbf{Basic Models (Group A)}. These methods use RNN (e.g., LSTM, GRU) or GNN (e.g., GCN, GAT) to encode the problem text into a feature vector and decompose this vector into an expression. We compare our method with various foundational methods without any language models: DNS \citep{wang2017DNS}, GTS \citep{Xie19Goal_driven}, T-RNN \citep{Wang2019template}, Group-ATT \citep{li2019modeling}, TSN-MD \citep{zhang2020teacher}, SAU-Solver \citep{Sem2020Qin}  and HMS \citep{HMS2021Lin}, Seq2Prog \citep{mathqa2019}, NumS2T \citep{wu2021math}, Graph2tree-Z \citep{ZhangGraph2Tree2020} and MWP-Teacher \citep{liang2021solving}.

    \item \textbf{PLM-enhanced Models (Group B)}. These methods mainly use pre-trained language models (e.g., BERT \citep{Bert}, RoBERTa \citep{RoBERTa2019}) as encoders and capture the general linguistic and semantic information. These baselines include UniLM \citep{Dong2019unified}, BERT-Tree \citep{li2022seeking}, MWP-BERT \citep{liang2022mwp}, mBERT-LSTM \citep{tan2021inves}, SUMC-Solver \citep{wang2022structure}, PseDual \citep{Bin2023solve}, MWP-NAS \citep{Bin2023non}, C-MWP \citep{Liang2023Compositional}, and PLM-enhanced variants of GTS and Graph2Tree-Z (e.g., Ro-GTS and Ro-Graph2Tree-Z).

    \item \textbf{LLM Models (Group C)}. Recently, methods using large models have demonstrated the remarkable reasoning ability for MWP solving. We compare our DivKD with methods using large models listed as follows: T5 \citep{Raffel2020T5}, GPT-3.5-turbo \citep{GPT-3}, LLaMA \citep{Touvron2023llama}, Math-PUMA \citep{mathpuma2024} and Math-LLaVA\footnote{Math-LLaVA is a multimodal LLM designed to exploit visual information to enhance the mathematical reasoning capabilities of multimodal. The comparison of Math-LLaVA is not provided since Math-LLaVA uses picture resources as essential inputs, making it difficult to reproduce their results. We leave these potential improvements in the future work.} \citep{mathllava2024}. We also consider the impact that using more powerful language models (e.g., LLaMA) as teacher models, denoted LLaMA-GTS + DivKD and LLaMA-Graph2Tree-Z + DivKD, which use the LLaMA as the input text encoder.

\end{itemize}

\begin{table}[htbp]
  \centering
  \caption{Answer accuracy on Math23K and MAWPS datasets: Note that the results evaluated on the public test set are denoted as ``Math23K'', and the results using 5-fold cross-validation are denoted as ``Math23K*'' and ``MAWPS*''. The symbol ``$^{\clubsuit}$'' denotes our re-produced results with the public released model.}
    \begin{tabular}{c|l|c|ccc}
    \toprule
    Category & Model & Size  & Math23K & Math23K* & MAWPS* \\
    \hline
    \multirow{10}[2]{*}{{Group A}} & DNS   & 4M    & 58.1  & -     & 59.5 \\
          & T-RNN & 8M    & 66.9  & 65.2  & 66.8 \\
          & Group-ATT & 9M    & 69.5  & 66.9  & 76.1 \\
          & GTS   & 14M   & 75.6  & 74.3  & 82.6 \\
          & Graph2Tree-Z & 16M   & 77.4  & 75.5  & 83.7 \\
          & TSN-MD & 22M   & 77.4  & 75.1  & 84.4 \\
          & SAU-Solver & 21M   & 76.2  & 74.8  & 75.5 \\
          & HMS   & 19M   & 76.1  & -     & 80.3 \\
          & NumS2T & 24M   & 78.1  & 75.9  & 84.3 \\
          & MWP-Teacher & 32M   & 79.1  & 77.2  & 84.2 \\
    \hline
    \multirow{8}[2]{*}{{Group B}} & UniLM & 340M  & 77.5  & -     & 78.0 \\
          & MWP-BERT & 130M  & 84.7  & 82.4  & - \\
          & Ro-GTS & 135M  & 84.0  & 82.0  & 87.1 \\
          & Ro-Graph2Tree-Z & 137M  & 84.9  & 82.4  & 88.7 \\
          & SUMC-Solver & 135M  & 77.4  & -     & 79.9 \\
          & C-MWP & 144M  & 86.1  & 84.3  & 89.1 \\
          & PseDual(GCN) & 128M  & 84.6  & -     & 92.4 \\
          & MWP-NAS & 121M  & 86.1  & -     & 91.4 \\
    \hline
    \multirow{6}[2]{*}{{Group C}} & T5    & 3B    & 63.2  & -     & 70.4 \\
          & GPT-3.5-turbo & 175B  & 54.8  & -     & 91.0 \\
          & LLaMA            & 7B    & 66.1  & -     & 90.1 \\
          & MATH-PUMA       & 7B    & 78.5$^{\clubsuit}$ & -     & 82.8$^{\clubsuit}$ \\
          & LLaMA-GTS          & 7B    & 84.8  & 82.9  & 90.6 \\
          & LLaMA-Graph2Tree-Z  & 7B   & 85.5 & 84.7  & 91.3 \\
    \hline
    \multirow{6}[2]{*}{{Ours}} 
          & GTS + DivKD              & 14M   & 77.7 (↑ 2.1) & 75.8 (↑ 1.5)  & 86.2 (↑ 3.6)  \\
          & Graph2Tree-Z + DivKD     & 16M   & 80.2 (↑ 2.8)  & 77.6 (↑ 2.1)  & 86.7 (↑ 4.0) \\
          & Ro-GTS + DivKD           & 135M  & 85.1 (↑ 1.1)  & 83.3 (↑ 1.3)  & 88.6 (↑ 1.5) \\
          & Ro-Graph2Tree-Z + DivKD  & 137M  & 86.2 (↑ 1.3)  & 84.5 (↑ 2.1)  & 90.1 (↑ 1.4)  \\
          & LLaMA-GTS + DivKD        & 7B    & 85.7 (↑ 0.9)  & 83.6 (↑ 0.7) & 91.3 (↑ 0.7) \\
          & \textbf{LLaMA-Graph2Tree-Z + DivKD} & 7B  & \textbf{86.7} (↑ 1.2) & \textbf{85.6} (↑ 0.9)  & \textbf{92.8} (↑ 1.5)\\
    \bottomrule
    \end{tabular}%
  \label{tab_res}%
\end{table}%

\begin{table}[htbp]
  \centering
  \caption{Results of various models on MathQA and SVAMP benchmarks: $\dagger$Results are from \citep{Bin2023solve}. The symbol ``$^{\clubsuit}$'' denotes our re-produced results with the public released model.}
    \begin{tabular}{l|l|c|cc}
    \toprule
    Category & Model & Size  & MathQA & SVAMP \\
    \hline
    \multirow{4}[2]{*}{{Group A}}
          & DNS        & 4M     & 65.7   & 18.7 \\
          & Group-ATT   & 9M    & 70.4  & 21.5 \\
          & Seq2Prog    & 10M   & 57.2  & - \\
          & NumS2T      & 24M   & 72.7  & 37.0 \\
    \hline
    \multirow{4}[1]{*}{{Group B}} 
          & BERT-Tree        & 121M  & 75.1 & 32.4 \\
          & mBERT-LSTM       & 132M  & 74.7 & - \\
          & PseDual(GCN)     & 128M  & 78.9 & - \\
          & Ro-GTS           & 135M  & 73.5 & 41.0 \\
          & Ro-Graph2Tree-Z  & 137M  & 74.4 & 43.8 \\
    \hline
    \multirow{4}[2]{*}{{Group C}} 
          & GPT-3.5-turbo       & 175B    & 73.5               & \textbf{69.3}$^{\dagger}$ \\
          & LLaMA               & 7B      & 75.7               & 38.0  \\
          & Math-PUMA           & 7B      & 54.3$^{\clubsuit}$ & 68.2$^{\clubsuit}$ \\
          & LLaMA-Graph2Tree-Z  & 7B      & 76.2               & 51.3 \\
    \hline
    \multirow{2}[2]{*}{Ours} 
          & Ro-GTS + DivKD                  & 135M  & 77.1 (↑ 3.6)           & 42.8 (↑ 1.8)\\
          & Ro-Graph2Tree-Z + DivKD         & 137M  & 78.7 (↑ 4.3)           & 45.3 (↑ 1.5)\\
          & {LLaMA-Graph2Tree-Z + DivKD}    & 7B    & \textbf{79.3} (↑ 3.1)  & 52.5 (↑ 1.2)\\
    \bottomrule
    \end{tabular}%
  \label{tab_res_mathqa}%
\end{table}%

\subsection{Main Results (RQ1 \& RQ2)}

Table~\ref{tab_res} and Table \ref{tab_res_mathqa} present the experimental results on Math23K, MAWPS, MathQA and SVAMP datasets. We use GTS, Graph2Tree-Z, and their language model variants (e.g., Ro-Graph2Tree-Z, LLaMA-Graph2Tree-Z) as the foundational models within the teacher-student framework, and apply the proposed DivKD method to these basic models. We compare our method against strong baselines and analyze each group.

From Tabel \ref{tab_res} and Table \ref{tab_res_mathqa}, we derive several observations:
(1) The models using the graph encoder significantly outperform these models using the sequence encoder (e.g., GTS and Group-ATT vs. Graph2Tree-Z and NumS2T). The reason is that GNN can capture global and long-distance relations in the problem texts, providing more semantic information for decoding.
(2) According to the comparison in Group B (e.g., Ro-Graph2Tree-Z vs. MAW-NAS), considering the structure information of expression can significantly improve the performance due to tree structures can decrease the search space and increase the efficiency of decoder.
(3) From the results in Group A and Group B, we can find that numerical knowledge can help math operation between numbers, such as UniLM vs. MWP-BERT, achieve better results on answer accuracy metrics. 
(4) The language model makes it easier to solve MWPs automatically. From Group B and Group A, we observe that models using PLMs achieve significant improvements in solving MWP (e.g., Graph2Tree-Z vs. Ro-Graph2Tree-Z). In particular, we replace the PLMs (e.g., RoBERTa-base) of Graph2Tree-Z with the more powerful language models (e.g., LLaMA), the LLaMA-Graph2Tree-Z achieve the improvement of 2.6\% and 1.8\% on MAWPS and MathQA.
(5) Compared to existing methods, our method learns diverse and high-quality information from the teacher model and sample latent variables, achieving 86.7\%, 92.8\% and 79.3\% on Math23K, 5-fold MAWPS and MathQA. Furthermore, we observe that when using GTS as the basic teacher and student models, the proposed ``GTS + DivKD'' outperforms the GTS model by 2.1\% (77.7\% vs. 75.6\%) on Math23K, 1.5\% (75.8\% vs. 74.3\%) on 5-fold Math23K, and 3.6\% (86.2\% vs. 82.6\%) on 5-fold MAWPS. When taking Graph2Tree-Z as the base model, our method Graph2Tree-Z + DivKD also achieves better accuracy than the original Graph2Tree-Z across all datasets. These experimental results demonstrate that our proposed DivKD method can effectively improve performance by modelling the diversity of solutions via diversity prior and adaptive diversity KD.


Table \ref{tab_res_mathqa} presents the accuracy comparison on the more challenging benchmark datasets, MathQA and SVAMP. We can find that these basic models using our DivKD method achieve significant improvements. For instance, our solver ``Ro-Graph2Tree-Z + DivKD'' surpasses Ro-Graph2Tree-Z by 4.3\% (78.7\% vs. 74.4\%) on MathQA, highlighting the effectiveness of our approach in enhancing problem-solving capabilities. We also analyze the improvements, and the results further confirm that our DivKD approach generate diverse and correct solutions, while the baselines are limited by their single solution approach. On the SVAMP dataset, our approach achieves an answer accuracy of 52.5\%, which is lower than GPT-3.5-turbo and Math-PUMA, but higher than other baseline methods. We observe that the average number of operators in the SVAMP is 1.33, significantly lower than in MathQA (average of 4.63) and Math23K (average of 2.70), indicating that SVAMP solutions are less complex. This complexity disparity explains why large language models (LLMs) perform well on SVAMP. In contrast, our DivKD approach is designed to model solution diversity more effectively, which is crucial for addressing more complex problem sets.

Among the baselines, TSN-MD is a related model that also uses the KD method with GTS as basic teacher and student models. From Table~\ref{tab_res}, we can see that our method ``GTS + DivKD'' outperforms TSN-MD on Math23K, 5-fold Math23K and 5-fold MAWPS (e.g., 77.7\% vs. 77.4\% on Math23k, 75.8\% vs. 75.1\% on 5-fold Math23K and 86.2\% vs. 84.4\% on 5-fold MAWPS).  TSN-MD simply performs distillation using soft labels from the teacher model without considering the quality of these soft labels, which could result in learning misinformation due to noisy labels. Instead, our method is able to adaptively select high-quality labels for KD, thereby mitigating the impact of noisy soft labels generated from the teacher model. Additionally,  TSN-MD uses multiple decoders in the student model to generate diverse solution equations, which is time-consuming. In contrast, we propose a diversity prior to efficiently modeling the diversity of solution equations using only one decoder, without increasing the complexity of the original GTS model. Therefore, our proposed DivKD method is superior to the previous KD-based method (e.g., TSN-MD) in both performance and efficiency.

\begin{table}
\caption{Ablation Study. }
\centering
\begin{tabular}{c|c|c}
\hline
Model& Math23K & MAWPS\\
\hline
GTS  &75.6 &82.6 \\
GTS + DivKD &77.7 &86.2 \\
\hline
GTS + DivKD w/o S\&H &76.4 &84.6 \\
GTS + DivKD w/o H&76.9 &85.0\\
GTS + DivKD w/o S&77.2 &85.9 \\
GTS + CVAE + SoftKD&76.6 &84.7 \\
\hline

\hline
Graph2Tree-Z &77.4 &83.7 \\ 
Graph2Tree-Z + DivKD &80.2 &86.7 \\
\hline
Graph2Tree-Z + DivKD w/o S\&H &78.2 &85.1 \\
Graph2Tree-Z + DivKD w/o H&78.9 &85.5\\
Graph2Tree-Z + DivKD w/o S&79.5 &86.1 \\
Graph2Tree-Z + CVAE + SoftKD  &78.4 &85.2 \\
\hline
\end{tabular}
\label{tab_ablation}
\end{table}

\begin{figure*}[]
\centering
\subfigure[Math23K]{
\begin{minipage}[t]{0.5\linewidth}
\centering
\includegraphics[width=0.99\textwidth]{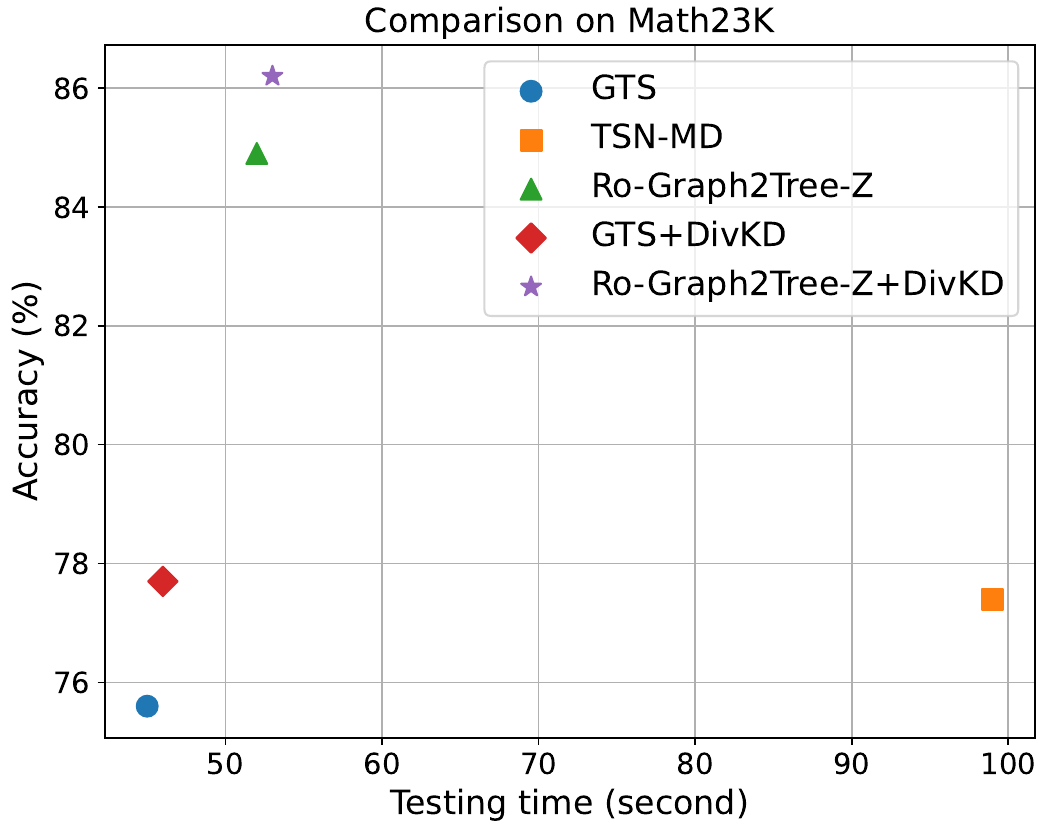}
\end{minipage}%
}%
\subfigure[MAWPS]{
\begin{minipage}[t]{0.5\linewidth}
\centering
\includegraphics[width=0.99\textwidth]{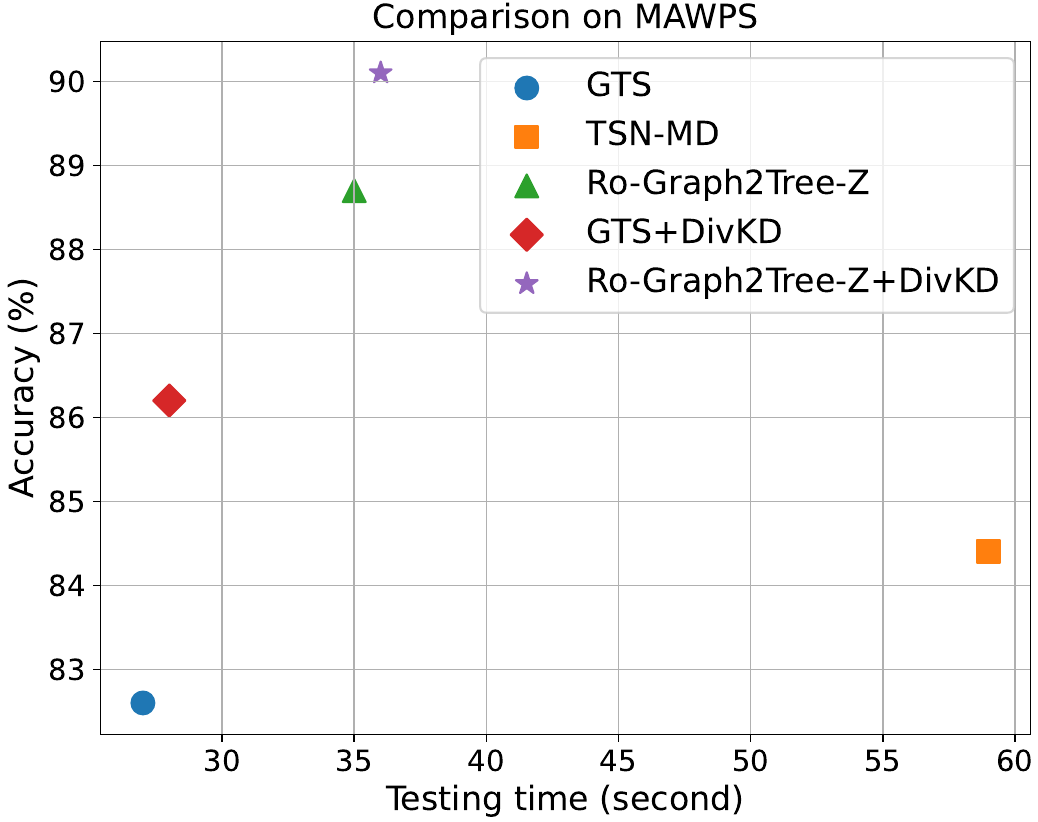}
\end{minipage}%
}%
\caption{Comparison of testing time between baselines and our proposed methods (e.g., GTS+DivKD and Ro-Graph2TreeZ+DivKD) for Math23K and MAWPS datasets.}
\label{fig_time}
\end{figure*}

\subsection{Ablation Study (RQ3)}
To better understand the impact of different components in the proposed DivKD method, we conduct ablation studies by constructing some variants of DivKD. As shown in Table~\ref{tab_ablation}, we use GTS and Graph2Tree-Z as basic teacher and student models and design four variants with different KD strategies:
\begin{itemize}
    \item ``GTS/Graph2Tree-Z + DivKD w/o S\&H'' denotes the GTS/Graph2Tree-Z model enhanced with diversity prior using CVAE, but without using adaptive soft and hard KD method.
    \item ``GTS/Graph2Tree-Z + DivKD w/o H'' only uses  AdaSoftKD to train the student model without using AdaHardKD.
    \item ``GTS/Graph2Tree-Z + DivKD w/o S'' only uses  AdaHardKD to train the student model without using AdaSoftKD. 
    \item ``GTS/Graph2Tree-Z + CVAE + SoftKD'' denotes using a conventional soft KD method to train the student model, without considering the quality of soft labels. 
\end{itemize}

As shown in Table~\ref{tab_ablation}, the proposed DivKD model without using KD outperforms the basic GTS/Graph2Tree-Z models on both datasets. For example, on MAWPS dataset, the ``GTS + DivKD w/o S\&H'' achieves an accuracy of 84.6\%, surpassing the performance of GTS (e.g., 82.6\%), and ``Graph2Tree-Z + DivKD w/o S\&H'' also outperforms original Graph2Tree-Z by a margin of 1.4\% (e.g., 85.1\% vs. 83.7\%).  This reveals that our DivKD model can effectively learn the diversity of solution equations by using diversity prior enhanced model.  

In addition, by applying KD, the performance can be further improved. The ``GTS/Graph2Tree-Z +DivKD w/o H'' models, which adaptively distill knowledge from high-quality  soft labels, obtain better performance than ``GTS/Graph2Tree-Z + CVAE + SoftKD'', indicating the effectiveness of the proposed adaptive KD method. Furthermore, we can observe that ``GTS/Graph2Tree-Z + DivKD w/o S'' performs better than other KD methods (e.g., conventional soft KD and adaptive soft KD), showing that the adaptive hard KD is superior to other KD methods since we can directly evaluate the answer of hard equations to filter out noise labels. Using both the adaptive hard KD and adaptive soft KD, our ``GTS + DivKD'' can achieve the best performance. In summary, the ablation studies demonstrate that all the components in the proposed DivKD model play important roles in the improvement of performance.

\subsection{Efficiency Analysis (RQ4)}
In this section, we compare the efficiency of the proposed model with existing baseline models, including GTS, TSN-MD and Ro-Graph2Tree-Z. GTS serves as the foundational model structure for both TSN-MD and our proposed ``GTS+DivKD''. TSN-MD is a related model that aims to model the diversity of solution equations using multiple decoders. Ro-Graph2Tree-Z and ``Ro-Graph2Tree-Z+DivKD'' utilize pre-trained Roberta model and graph structure to enhance the encoding of input problem. We evaluate these models on the test sets of Math23K and MAWPS and present their testing times in Figure~\ref{fig_time}. All experiments are conducted on a single RTX A6000 48G graphics card to ensure a fair comparison. 

Comparing among GTS, TSN-MD and ``GTS+DivKD'', we observe that the running time of ``GTS+DivKD'' is similar to that of the basic GTS model, whereas TSN-MD requires significantly more time for inference than both GTS and ``GTS+DivKD''. This increased inference time in TSN-MD is due to its use of two decoders to generate solution equations. In contrast, ``GTS+DivKD'' utilizes only one decoder and incorporates a CVAE to better model the diversity of solution equations, thereby enhancing performance without compromising efficiency. Furthermore, by employing PLMs, Ro-Graph2Tree-Z and ``Ro-Graph2Tree-Z+DivKD'' improve performance on both Math23K and MAWPS datasets with only a slight increase in computational time compared to GTS and ``GTS+DivKD''. But the testing time of Ro-Graph2Tree-Z and ``Ro-Graph2Tree-Z+DivKD'' are still much less than TSN-MD model. This is because the tree-based decoding process is very time-consuming, and the TSN-MD model, which uses multiple decoders, will inevitably significantly increase the testing time.
Therefore, using multiple decoders to improve the performance is not a good choice. Different from TSN-MD, our DivKD method can be conveniently plugged into different models and can improve model effectiveness without causing an increase in inference time. This renders our approach feasible for practical application.

\begin{figure*}[]
\centering
\subfigure[Diversity analysis on Math23K]{
\begin{minipage}[t]{0.46\linewidth}
\centering
\includegraphics[width=0.99\textwidth]{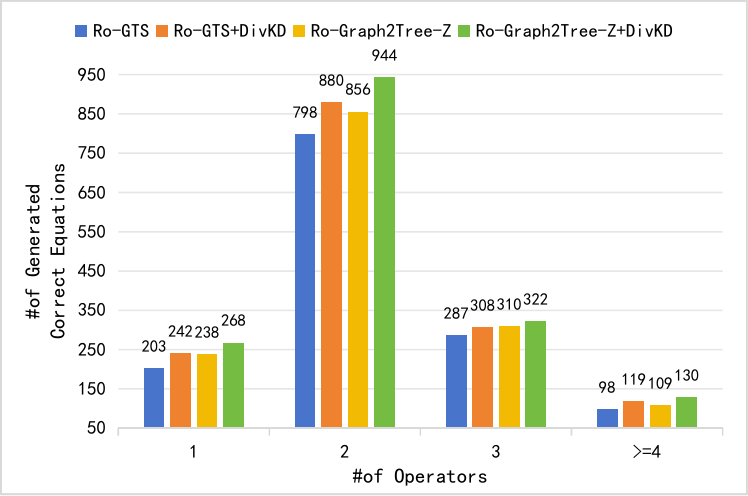}
\end{minipage}%
}%
\subfigure[Diversity analysis on MathQA]{
\begin{minipage}[t]{0.49\linewidth}
\centering
\includegraphics[width=0.99\textwidth]{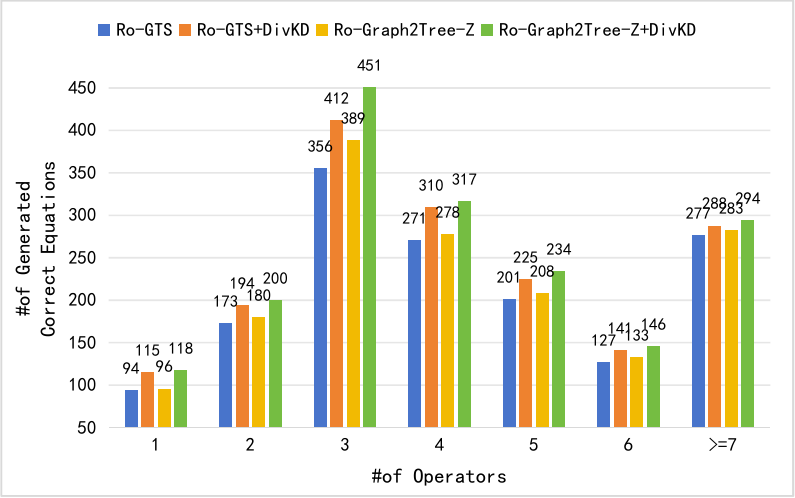}
\end{minipage}%
}%
\caption{A quantitative analysis of generated correct expressions between basic models (e.g., Ro-GTS) and the student models (e.g., Ro-GTS+DivKD) on Math23K and MathQA datasets.}
\label{fig_analysis}
\end{figure*}

\subsection{Quantitative Analysis (RQ5)}

In this section, we conduct a quantitative analysis to count the number of the correct solution equations generated by the teacher models (e.g., Ro-GTS and Ro-Graph2Tree-Z) and student models (e.g., ``Ro-GTS + DivKD'' and ``Ro-Graph2Tree-Z + DivKD'') in Figure \ref{fig_analysis}. On the Math23K and MathQA datasets, we can see that the student models produce a significantly higher number of correct equations compared to the teacher models. For instance, on the subset of MathQA with 3 operators, our ``Ro-Graph2Tree-Z+DivkD'' outperforms the Ro-Graph2Tree-Z by 62. This improvement is attributed to DivKD's ability to learn the diversity distribution of solution equations, thereby generating a wider range of potential solutions during the test phase. Instead, Ro-GTS and Ro-Graph2Tree-Z are limited to learning only one ground-truth solution equation provided for each math problem in the benchmark datasets. Overall, these results demonstrate that our method enhances both the diversity and correctness of generated equations, showcasing DivKD's capability to produce more accurate and varied solutions.

\subsection{Case Study (RQ5)}

\begin{figure}[!t]
\centering
\includegraphics[width=0.9\textwidth]{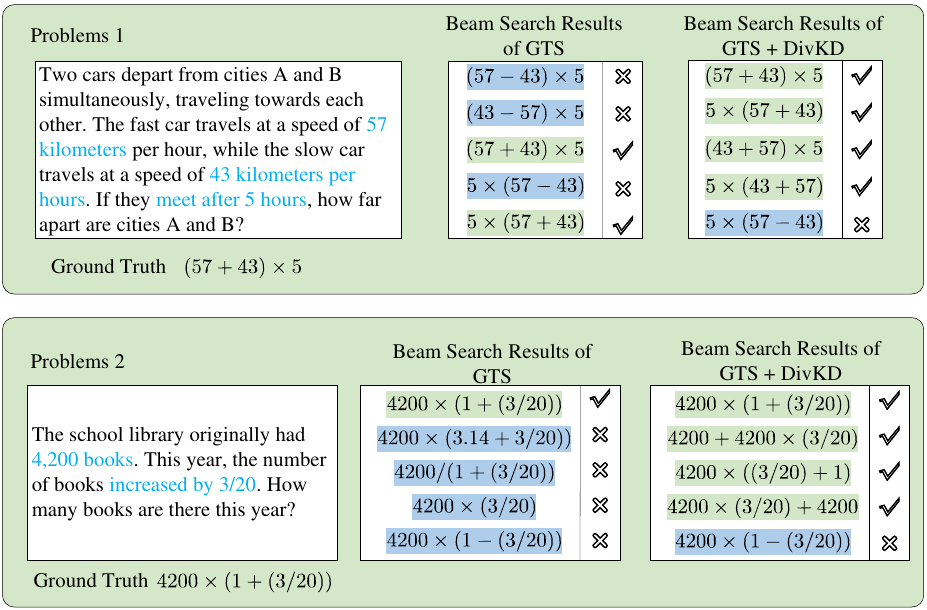}
\caption{Two examples of the generated results on Math23K using GTS and the proposed GTS+DivKD.}
\label{fig_case}
\end{figure}

To better understand the superiority of the proposed method, we conduct a case study by comparing the results generated by the original GTS and our proposed ``GTS+DivKD''. As shown in Figure~\ref{fig_case}, for Problem 1, the original GTS model generates a wrong solution equation at the first rank of the five beam search results and only two correct solution equations are ranked at the top five results.  
Instead, our proposed ``GTS+DivKD'' can correctly predict four solution equations and rank them at top positions. In the second example, although the GTS predicts a correct solution equation ranking at the first position, it fails to model the diversity of solution equations since only one correct equation is ranked in the top 5 results. Different from the original GTS, our proposed ``GTS+DivKD'' can generate diverse correct solution equations in the beam search results. 
These examples demonstrate that the proposed DivKD method is able to model the diversity distribution of multiple solution equations for MWPs.

\subsection{Discussion}
In recent years, LLMs and LMMs such as GPT-3.5-turbo \citep{GPT-3} and Math-PUMA \citep{mathpuma2024} have made remarkable strides in solving MWPs with the help of chain-of-thought (CoT) prompting \citep{Chen2022parogram, Wang2023self,Xie2024adversarial}. Although our model does not outperform these advanced large-scale models on the MWP task, we believe our work contributes significant value to the research field. Real-time or resource-constrained is crucial for many real-world applications, such as smart education. However, existing LLMs have hundreds of millions of parameters and exorbitant API, resulting in computational inefficiency and impracticality for real-world situations. In contrast, our method achieves competitive performance compared to LLMs while offering faster response times and requiring fewer parameters for the MWP task. Furthermore, our method is a natural solution that distills the knowledge from a bigger model into smaller, more efficient student models, which can potentially mitigate the issues of LLMs for practical situations. Objectively speaking, our gold is designing practical MWP solvers to balance effectiveness and efficiency.

\section{Conclusion and Future Work}
\label{conclusion}
In this paper, introduce a novel Diversity-Enhanced Knowledge Distillation (DivKD) model for solving Math Word Problems (MWPs). Our approach effectively models diverse solution equations in MWPs and extracts high-quality knowledge from the teacher model. Specifically, we enhance the student model with a diversity prior by integrating a Conditional Variational Autoencoder (CVAE) module into conventional encoder-decoder architectures (e.g., GTS, Graph2Tree-Z, and Ro-Graph2Tree-Z). This integration enables the model to capture the diversity of equations. Moreover, to overcome the limited annotation equations in datasets, we propose an Adaptive Diversity Knowledge Distillation (AdaKD) method that selects high-quality knowledge from the teacher model, providing an additional supervisory signal to guide the student model's learning process. Empirical evaluations on four widely used benchmark datasets demonstrate the superiority of the proposed DivKD model in MWP.

Future research can proceed in two primary directions. 
First, despite the strong performance of large-scale models such as GPT-3.5-turbo and LLaMA in MWP solving, their practical deployment remains limited due to their substantial model sizes and high computational costs. 
Online educational applications require models that can serve millions of students with low latency while designing personalized learning paths, presenting significant challenges in both effectiveness and efficiency. 
Therefore, a promising direction is to focus on compressing these large models into smaller, more efficient versions without compromising performance, thereby facilitating their integration into personalized mobile educational platforms.
Second, we aim to evaluate the robustness of the proposed DivKD method on more complex MWP datasets, including CM17K \citep{neural2021qin}, GSM8K \citep{Cobbe2021gsm8k}, and MATH \citep{Hendrycks2021math}. Assessing performance on these datasets will provide deeper insights into the model's ability to handle a wider variety of problem complexities and further validate its effectiveness.


\section*{Data Availability}
Data will be made available on request.

\section*{ACKNOWLEDGMENTS}
This work was supported by the National Natural Science Foundation of China under Grant 62377021, the China Postdoctoral Science Foundation under Grant Number 2024M751062, financially supported by self-determined research funds of CCNU from the colleges’ basic research and operation of MOE (No. CCNU24XJ010, CCNU22QN015 and CCNU24ai011), the Natural Science Foundation of Hubei Province for Distinguished Young Scholars (No. 2023AFA096), and the Wuhan Knowledge Innovation Project (No. 2022010801010278). This work was also supported by the Natural Sciences and Engineering Research Council (NSERC) of Canada, an NSERC CREATE award in ADERSIM3 and the York Research Chairs (YRC) program.




\bibliographystyle{cas-model2-names}

\bibliography{ref_math}





\end{CJK}
\end{document}